\documentclass[10pt,twocolumn,letterpaper]{article}

\usepackage{iccv}              
\usepackage{multirow}

\definecolor{iccvblue}{rgb}{0.21,0.49,0.74}
\usepackage[pagebackref,breaklinks,colorlinks,allcolors=iccvblue]{hyperref}

\def\paperID{10200} 
\def\confName{ICCV}
\def\confYear{2025}

\title{Grounding-MD: Grounded Video-language Pre-training\\ for Open-World Moment Detection}

\author{
Weijun Zhuang$^{1,2\ *}$ \quad
Qizhang Li$^{1,2\ *}$ \quad
Xin Li$^{2\ \dag}$ \quad
Ming Liu$^{1}$ \quad
Xiaopeng Hong$^{1}$ \\
Feng Gao$^{3}$ \quad
Fan Yang$^{3}$ \quad
Wangmeng Zuo$^{1\ \dag}$ \\
\vspace{-0.5em}
\\
$^1$Harbin Institute of Technology \quad
$^2$Pengcheng Laboratory \quad
$^3$Peking University \\
}

\begin{document}
\maketitle
\begin{abstract}

Temporal Action Detection and Moment Retrieval constitute two pivotal tasks in video understanding, focusing on precisely localizing temporal segments corresponding to specific actions or events. 
%
%
Recent advancements introduced Moment Detection to unify these two tasks, yet existing approaches remain confined to closed-set scenarios, limiting their applicability in open-world contexts. 
To bridge this gap, we present Grounding-MD, an innovative, grounded video-language pre-training framework tailored for open-world moment detection.
Our framework incorporates an arbitrary number of open-ended natural language queries through a structured prompt mechanism, enabling flexible and scalable moment detection. 
Grounding-MD leverages a Cross-Modality Fusion Encoder and a Text-Guided Fusion Decoder to facilitate comprehensive video-text alignment and enable effective cross-task collaboration. 
Through large-scale pre-training on temporal action detection and moment retrieval datasets, Grounding-MD demonstrates exceptional semantic representation learning capabilities, effectively handling diverse and complex query conditions. 
%
Comprehensive evaluations across four benchmark datasets including ActivityNet, THUMOS14, ActivityNet-Captions, and Charades-STA demonstrate that Grounding-MD establishes new state-of-the-art performance in zero-shot and supervised settings in open-world moment detection scenarios.
%
All source code and trained models will be released.

\end{abstract}    
\section{Introduction}
\label{sec:introduction}

Video understanding \cite{li2024mvbench,lin2019tsm,fan2018end} has emerged as a pivotal research area in computer vision, driven by the exponential growth of large-scale video datasets \cite{wang2023internvid,bain2021frozen} and the widespread reliance on video streaming platforms for information consumption. A critical challenge in this domain is the precise identification and localization of actions or events within video sequences, which is essential for enabling machines to interpret and reason about temporal dynamics. Two prominent tasks addressing this challenge are temporal action detection and moment retrieval.

Temporal Action Detection (TAD) \cite{zhang2022actionformer,liu2024end,shi2023tridet} refers to the identification of specific actions and the precise localization of their temporal boundaries within video sequences. In parallel, Moment Retrieval (MR) \cite{zeng2024unimd,qian2024momentor} involves retrieving temporally localized video segments based on natural language descriptions of actions or events. The primary distinction between TAD and MR lies in their input modalities. While TAD operates within a closed-set paradigm using predefined action labels or ground truth annotations, MR adopts an open-ended approach by processing natural language queries that describe target moments. Despite this divergence in input representation, both tasks converge on a fundamental objective: the comprehensive understanding of temporal context and video content semantics, coupled with the accurate temporal localization of action-specific moments within video content.

\begin{figure}[t]
  \centering
  \includegraphics[width=0.95\linewidth]{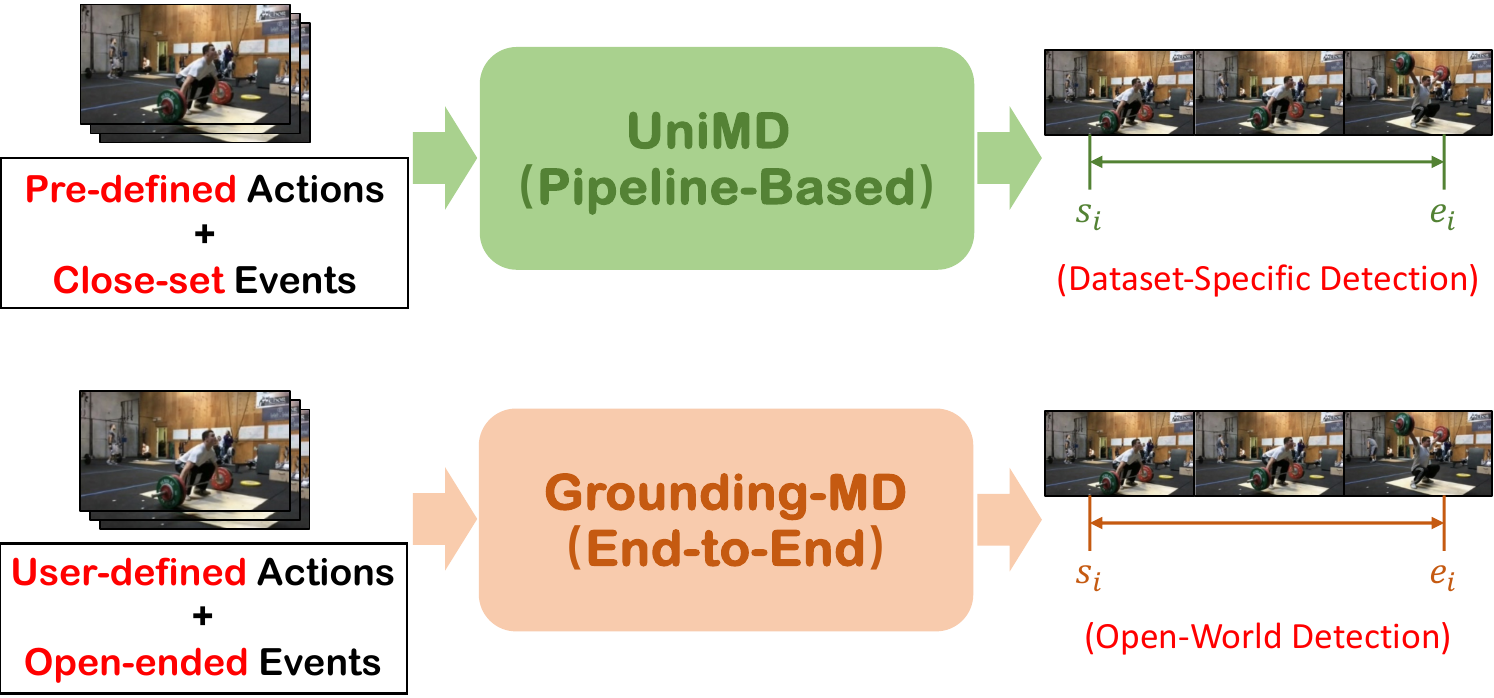}
  \caption{\textbf{Illustration of UniMD \cite{zeng2024unimd} and Grounding-MD.} UniMD operates under a closed-set assumption, limiting its applicability in open-world scenarios. In contrast, Grounding-MD supports user-defined action categories and open-ended natural language event descriptions, enabling it to adapt to diverse and dynamic user queries in open-world environments.}
  \label{fig:moti}
  \vspace{-1em}
\end{figure}

\begin{figure*}[tb]
  \centering
  \begin{subfigure}{0.48\linewidth}
    \centering
    \includegraphics[width=1.0\linewidth]{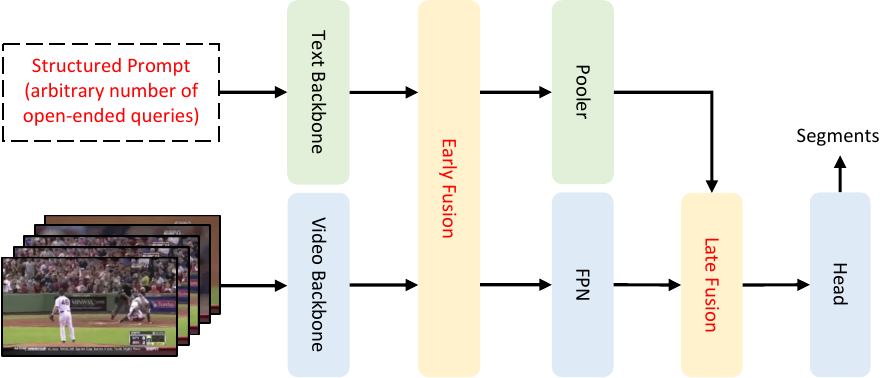}
    \caption{Deep Video-Text Understanding and Robust Cross-Task Collaboration}
    \label{fig:adv12}
  \end{subfigure}
  \hspace{8.0pt}
  \begin{subfigure}{0.48\linewidth}
    \centering
    \includegraphics[width=1.0\linewidth]{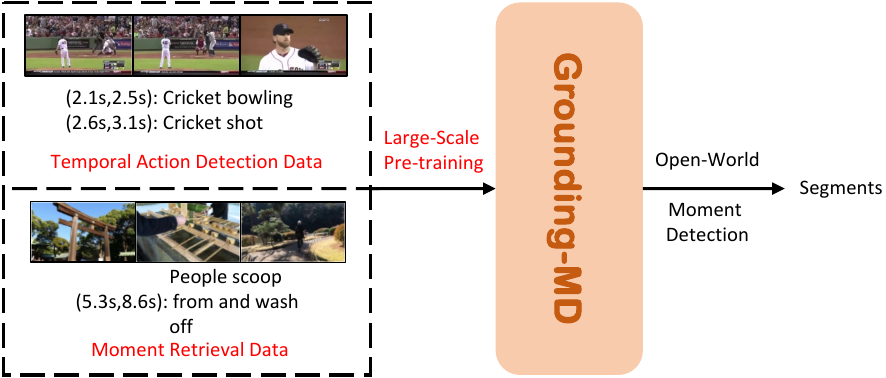}
    \caption{Comprehensive Video-Text Cognition}
    \label{fig:adv3}
  \end{subfigure}
  \caption{\textbf{Advantages of Grounding-MD.} (a) The early and late fusion strategies achieve optimal video-text alignment, enabling the model to gain a deeper understanding of the video-text data. Moreover, the structured prompt design allows the model to handle an arbitrary number of open-ended natural language queries, demonstrating excellent cross-task collaboration; (b) Through pre-training on large-scale temporal action detection and moment retrieval datasets, Grounding-MD develops robust semantic representation learning and video-text cognition abilities, enabling it to handle more complex and diverse query inputs.}
\end{figure*}

Despite their foundational similarity, TAD and MR have traditionally been approached as distinct tasks within the research community. While this segregated approach has driven significant advancements in both domains, it has largely neglected the potential collaboration between these complementary tasks. Recently, Zeng et al. \cite{zeng2024unimd} introduced Moment Detection as a unified task that bridges TAD and MR and proposed UniMD as a versatile model capable of simultaneously addressing moment detection. Through an innovative, collaborative training strategy, UniMD demonstrated that task-specific performance could be enhanced via cross-task interaction, achieving improved results across benchmarks. However, the applicability of UniMD remains constrained to closed-set scenarios with predefined datasets (Figure \ref{fig:moti}), limiting its ability to generalize to open-world environments characterized by diverse and dynamic user queries coupled with heterogeneous video content. This limitation underscores the need for more adaptable frameworks capable of handling the inherent complexity and variability of real-world applications.

To address the aforementioned issues, we present Grounding-MD, a novel large-scale video-text pre-training framework for open-world moment detection. Our approach introduces a unified formulation that integrates an arbitrary number of open-ended natural language queries through a structured prompt mechanism, serving as the conditional input for moment detection. This formulation fundamentally extends beyond UniMD \cite{zeng2024unimd} by supporting unrestricted query types and quantities, enabling flexible adaptation to diverse open-world scenarios. Grounding-MD embodies this formulation through three key innovations: 1) \textbf{Deep Video-Text Understanding} (Figure \ref{fig:adv12}). Grounding-MD incorporates a Cross-Modality Fusion Encoder and a Text-Guided Fusion Decoder to facilitate comprehensive early and late fusion of video and text modalities. This hierarchical fusion mechanism enables optimal cross-modal alignment, enhancing the ability to capture nuanced semantic relationships between visual and textual representations. 2) \textbf{Robust Cross-Task Collaboration} (Figure \ref{fig:adv12}). While our unified formulation supports the simultaneous handling of TAD and MR tasks, the inherent variability in text length between action categories and event descriptions results in training instability. Grounding-MD addresses this challenge through a novel Query-Wise Pooler that effectively balances structural biases across different query types, promoting more stable and effective cross-task collaboration. 3) \textbf{Comprehensive Video-Text Cognition} (Figure \ref{fig:adv3}). Grounding-MD leverages extensive pre-training on large-scale temporal action detection and moment retrieval datasets, establishing a robust foundation for semantic representation learning. This extensive pre-training endows the model with exceptional video-text cognitive capabilities, enabling superior performance in handling complex and diverse query conditions in open-world scenarios.

To validate the effectiveness of our proposed framework, we conducted experiments on four benchmark datasets: ActivityNet \cite{caba2015activitynet}, THUMOS14 \cite{idrees2017thumos}, ActivityNet-Captions \cite{krishna2017dense}, and Charades-STA \cite{gao2017tall}. Grounding-MD establishes new state-of-the-art performance in both zero-shot and supervised settings for open-world moment detection scenarios. Notably, despite utilizing a smaller model size and less pre-training data, Grounding-MD achieves favorable zero-shot performance compared to Video-LLM-based moment retrieval methods, highlighting its practical advantages in computational efficiency and scalability.

In this paper, we make the following contributions:
\begin{itemize}
\item We introduce Grounding-MD, a novel unified framework for open-world moment detection that integrates early-late fusion mechanisms with large-scale video-text pre-training. This framework achieves deep video-text understanding, robust cross-task collaboration, and comprehensive video-text cognition, establishing a new paradigm for open-world moment detection.
\item To the best of our knowledge, Grounding-MD is the first grounded video-language pre-training framework specifically designed for moment detection in open-world scenarios. Our work provides valuable insights and methodological foundations for future research in grounded video-language pre-training and moment detection.
\item We conduct extensive experiments under zero-shot and supervised settings for moment detection and demonstrate state-of-the-art
performance of Grounding-MD, underscoring its efficacy in
open-world moment detection.
\end{itemize}

\section{Related Work}
\label{sec:related_works}

\subsection{Moment Detection}

Moment Detection (MD) \cite{zeng2024unimd} encompasses two fundamental tasks: Temporal Action Detection (TAD) and Moment Retrieval (MR). TAD focuses on identifying the start and end times of specific actions within a continuous video sequence, along with classifying the nature of these actions. Existing methodologies in TAD can be broadly categorized into two approaches: two-stage methods \cite{zeng2019graph,bai2020boundary,xia2022learning,zhu2021enriching}, which first generate a set of action proposals and then refine their boundaries and classifications in a subsequent stage, and one-stage methods \cite{shi2023tridet,lin2017single,liu2024end,zhang2022actionformer}, which directly predict action categories and their temporal boundaries in a single pass, eliminating the need for proposal generation. In contrast, MR aims to locate specific moments or segments within a video based on a natural language query. This task necessitates a deep understanding of both visual content and textual semantics, requiring models to bridge the gap between video and language. MR methodologies can be divided into proposal-based \cite{anne2017localizing,yuan2019semantic,zhang2020learning} methods, which first generate candidate moments and then rank them based on their relevance to the query, and proposal-free \cite{liu2022umt,mun2020local,zeng2020dense} methods, which directly predict the most relevant moment in a single step, bypassing the proposal generation stage.

While both TAD and MR require an understanding of temporal context and video content to identify relevant action segments accurately, they have traditionally been treated as separate tasks. Recent work by Zeng et al. \cite{zeng2024unimd} introduced UniMD as a unified framework for TAD and MR. However, UniMD primarily operates in a closed-set setting with specific datasets, limiting its adaptability to open-world scenarios characterized by diverse and dynamic user queries and video content. Our work addresses these limitations by proposing a unified framework that bridges TAD and MR and effectively handles the complexities of open-world scenarios.

\subsection{Vision-Language Pre-training}

Visual-Language Pre-training has emerged as a cornerstone in multimodal learning, enabling models to understand and reason across visual and textual modalities. Pioneering works like CLIP \cite{radford2021learning} have demonstrated the effectiveness of large-scale noisy image-text pairs in learning robust image-level visual representations. Building on this foundation, open-world object detection methods \cite{li2022grounded,liu2024grounding,jiang2024t,yao2024detclipv3} have advanced spatial understanding by leveraging image labels, captions, and bounding boxes to develop region-level comprehension capabilities. Furthermore, several works \cite{wang2021actionclip,wang2023internvid,cheng2023vindlu} have extended vision-language pre-training techniques to video-language models. However, due to the high cost of acquiring fine-grained temporal annotations, most existing approaches focus on holistic video understanding rather than detailed moment-level analysis. Our work addresses this gap by exploring grounded video-language pre-training through integrating large-scale video-text datasets, thereby improving performance on both TAD and MR tasks in open-world scenarios.

\begin{figure*}[t]
  \centering
  \includegraphics[width=1.0\linewidth]{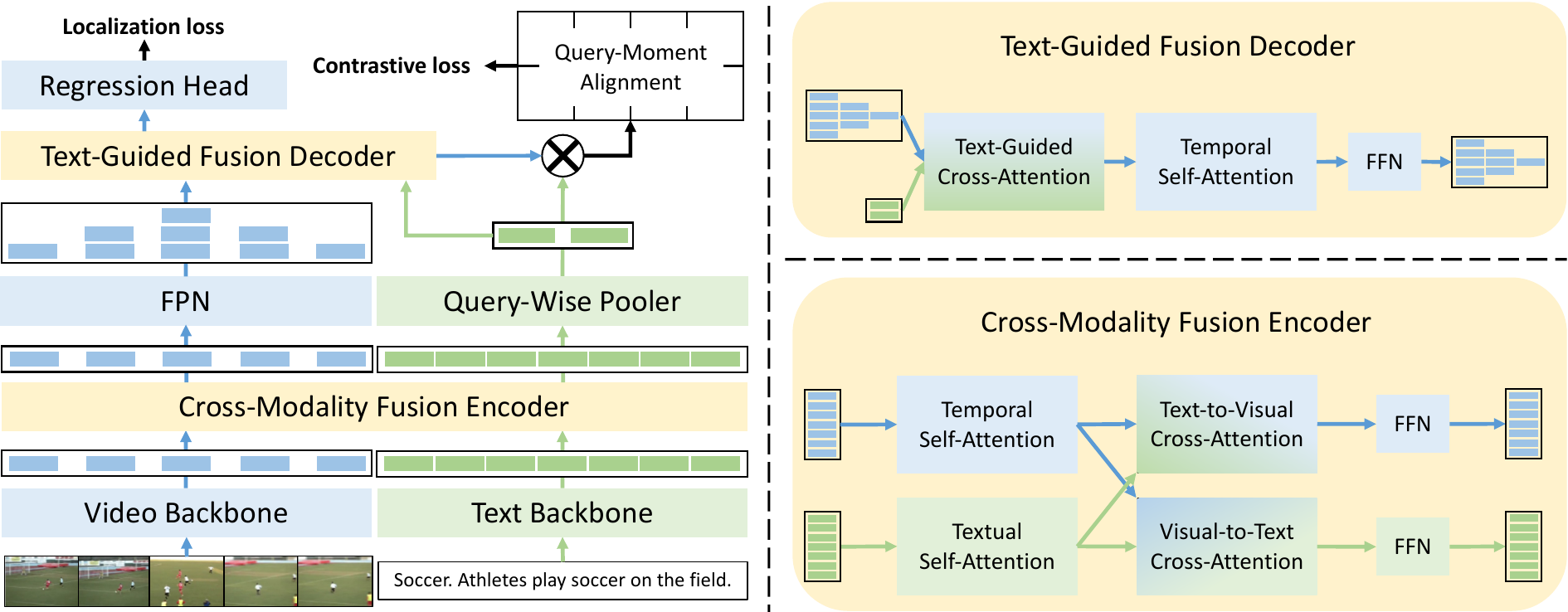}
  \caption{\textbf{Overview of the Grounding-MD framework.} The Cross-Modality Fusion Encoder performs early fusion of video and text features, establishing initial cross-modal alignment, while the Text-Guided Fusion Decoder conducts late fusion, refining the alignment through deeper interaction between modalities. Additionally, the Query-Wise Pooler addresses the training instability caused by the disparity in text lengths between action categories and event descriptions by generating balanced query-wise textual representations.}
  \label{fig:arch}
  \vspace{-1em}
\end{figure*}

\section{Method}
\label{sec:method}

This paper introduces a video-language pre-training pipeline designed to enhance the performance of open-world moment detection. In this section, we first present a unified formulation to integrate open-world TAD and MR (Section \ref{subsec:uf}). Next, we introduce Grounding-MD, our unified model that seamlessly incorporates the proposed formulation (Section \ref{subsec:ma}). Finally, we detail the data collection process for pre-training and the training methodology employed (Section \ref{subsec:to}).

\subsection{Unified Formulation}
\label{subsec:uf}

\noindent\textbf{Review of TAD and MR.} Given an untrimmed video $\mathbf{V}$, TAD focuses on identifying temporal segments corresponding to predefined action categories, providing both temporal localization and classification of actions.
\begin{equation}
\label{eq:tad}
    \{(seg,c_{tad})\}=f_{tad}(\mathbf{V}),
\end{equation}
where $seg=(s,e)$ represents the temporal segment for action category $c_{tad}$, with $s$ and $e$ denoting the start and end timestamps, respectively; $f_{tad}(\cdot)$ denotes a TAD method. In contrast, MR detects temporal segments corresponding to events described by natural language descriptions.
\begin{equation}
\label{eq:mr}
    \{seg\}=f_{mr}(\mathbf{V},E),
\end{equation}
where $E$ denotes the natural language descriptions of events, and $f_{mr}(\cdot)$ denotes a MR method.

Conceptually, TAD and MR share substantial similarities, as both tasks involve the retrieval of specific temporal segments from untrimmed videos. To explore the collaboration between these tasks, UniMD \cite{zeng2024unimd} introduces Moment Detection (MD), a unified formulation that bridges TAD and MR. While UniMD demonstrates the potential for cross-task collaboration, it is constrained to a closed-set setting tailored to specific datasets, limiting its adaptability to open-world scenarios where user queries and video content are more diverse and dynamic. This limitation underscores the need for a more flexible and scalable approach capable of handling the complexities of open-world applications.

\vspace{2mm}
\noindent\textbf{Unified Formulation for Open-World Moment Detection.} To address the gap in applying moment detection to open-world scenarios, we draw inspiration from recent advancements in open-world object detection \cite{li2022grounded,liu2024grounding} and propose a unified formulation for Open-World Moment Detection (OWMD). This formulation takes an untrimmed video $\mathbf{V}$ and a structured prompt $P$ as input and outputs the detected temporal segments.
\begin{equation}
\label{eq:md}
    \{seg\}=f_{owmd}(\mathbf{V},P),
\end{equation}
A structured prompt consists of an arbitrary number of open-ended natural language queries, where each query can represent any action category or event description.
\begin{equation}
\label{eq:prompt}
    P=\text{``query\_1. query\_2. ... . query\_n."}.
\end{equation}

\subsection{Model Architecture}
\label{subsec:ma}

We introduce Grounding-MD, a unified model designed for open-world moment detection. Grounding-MD takes an untrimmed video and a structured prompt as input and outputs detected temporal segments (Fig. \ref{fig:arch}), fully adhering to the open-world moment detection formulation we proposed.

Given an untrimmed video with $T$ frames $\mathbf{V}=\{\mathbf{v}_t\}_{t=1}^{T}$ and a structured prompt $P$ consisting of $L_q$ queries, we employ a video backbone and a text backbone to independently extract video features $\mathbf{F}_v{\in}\mathbb{R}^{T{\times}D}$ and text features $\mathbf{F}_p{\in}\mathbb{R}^{L_p{\times}D}$, where $T$ and $L_p$ denote the number of video frames and the number of tokens, respectively, and $D$ represents the feature dimension. A critical factor for success in the open-world setting is the effective alignment between vision and language. To achieve this, we propose two key components: a cross-modality fusion encoder and a text-guided fusion decoder, which facilitate early and late-stage deep fusion of video and text features. Both modules are built upon multi-head attention \cite{vaswani2017attention} $\mathbb{MHA}(Q,K,V)$, enabling comprehensive cross-modal interaction.

\begin{table*}[t]
    \caption{A detailed list of datasets used for pre-training. We train Grounding-MD using both temporal action detection and moment retrieval datasets, with a total of 357K instances.}
    \centering
    \begin{tabular}{cccccccc}
        \toprule
        Dataset & Category & Video & Instance & Duration & Domain & Main Task \\
        \midrule
        HACS \cite{zhao2019hacs} & 200 & 35K & 87K & 33.2s & action & temporal action detection \\
        FineAction \cite{liu2022fineaction} & 106 & 17K & 103K & 7.1s & action & temporal action detection \\
        InternVid-MR & - & 52K & 167K & 13.4s & open & moment retrieval \\
        \bottomrule
    \end{tabular}
\label{tab:datasets}
\end{table*}

\noindent\textbf{Cross-Modality Fusion Encoder.} Given that the video and text features are extracted through independent backbones, there is no inherent semantic connection between them. To establish effective cross-modal alignment, we introduce the Cross-Modality Fusion Encoder (CMFE), which performs an early fusion of video and text features. Specifically, CMFE takes the video features $\mathbf{F}_v$ and text features $\mathbf{F}_p$ as input and generates enhanced video and text features.
\begin{equation}
\label{eq:cmfe}
    \begin{aligned}
        &\mathbf{F}_v^{self}=\mathbb{MHA}(\mathbf{F}_v,\mathbf{F}_v,\mathbf{F}_v)+\mathbf{F}_v,\\
        &\mathbf{F}_p^{self}=\mathbb{MHA}(\mathbf{F}_p,\mathbf{F}_p,\mathbf{F}_p)+\mathbf{F}_p,\\
        &\mathbf{F}_v^{t2v}=FFN(\mathbb{MHA}(\mathbf{F}_v^{self},\mathbf{F}_p^{self},\mathbf{F}_p^{self})+\mathbf{F}_v^{self}),\\
        &\mathbf{F}_p^{v2t}=FFN(\mathbb{MHA}(\mathbf{F}_p^{self},\mathbf{F}_v^{self},\mathbf{F}_v^{self})+\mathbf{F}_p^{self}),
    \end{aligned}
\end{equation}
where $\mathbf{F}_v^{t2v}{\in}\mathbb{R}^{T{\times}D}$ and $\mathbf{F}_p^{v2t}{\in}\mathbb{R}^{L_p{\times}D}$ are the enhanced video and text features, respectively. This early fusion process enables the model to capture fine-grained interactions between visual and textual modalities, laying the foundation for robust cross-modal understanding. To further enhance the video features, we incorporate an FPN module, which supports multi-granularity temporal detection.
\begin{equation}
\label{eq:fpn}
    \begin{aligned}
        &\{\mathbf{F}_v^{fpn}\}=FPN(\mathbf{F}_v^{t2v}),\\
        &\mathbf{F}_v^{ms}=ConCat(\{\mathbf{F}_v^{fpn}\}),
    \end{aligned}
\end{equation}
where $ConCat(\cdot)$ is the concatenation operation; $\mathbf{F}_v^{ms}$ denote the multi-scale video features.

\noindent\textbf{Query-Wise Pooler.} The structural differences between action categories and natural language descriptions often result in significant variations in text length, with action categories typically being much shorter than natural language descriptions. This discrepancy biases the model towards optimizing MR task, leading to instability during joint training. To address this issue, we introduce the Query-Wise Pooler (QWP), which processes the enhanced text features to generate balanced query representations.
\begin{equation}
\label{eq:qwp}
    \begin{aligned}
        &\{\mathbf{F}_q^{qws}\}=QWS(\mathbf{F}_p^{v2t}),\\
        &\{\mathbf{F}_q^{pooled}\}=AvgPool(\{\mathbf{F}_q^{qws}\}),\\
        &\mathbf{F}_q^{qwp}=ConCat(\{\mathbf{F}_q^{pooled}\}),
    \end{aligned}
\end{equation}
where $QWS(\cdot)$ denotes query-wise split, which split text features into $L_q$ query-wise features $\{\mathbf{F}_q^{qws}\}$; $AvgPool(\cdot)$ is the average pooling operation; $\mathbf{F}_q^{qwp}{\in}\mathbb{R}^{L_q{\times}D}$ are the pooled query representations.

\vspace{2mm}
\noindent\textbf{Text-Guided Fusion Decoder.} After processing through FPN for video features and QWP for text features, both the structural and semantic information undergo certain transformations. To achieve deeper visual-text semantic alignment, we introduce the Text-Guided Fusion Decoder (TGFD), which performs a late-stage fusion of the refined video features and query representations.
\begin{equation}
\label{eq:tgfd}
    \begin{aligned}
        &\mathbf{F}_v^{tga}=\mathbb{MHA}(\mathbf{F}_v^{ms},\mathbf{F}_q^{qwp},\mathbf{F}_q^{qwp})+\mathbf{F}_v^{ms},\\
        &\mathbf{F}_v^{tgfd}=FFN(\mathbb{MHA}(\mathbf{F}_v^{tga},\mathbf{F}_v^{tga},\mathbf{F}_v^{tga})+\mathbf{F}_v^{tga}).
    \end{aligned}
\end{equation}

\subsection{Training Objective}
\label{subsec:to}

\noindent\textbf{Pre-training with Large-Scale Data.} To unify TAD and MR in open-world scenarios, our model requires extensive pre-training on large-scale, semantically rich data to achieve effective visual-text alignment. This alignment is crucial for enabling the model to perform precise moment detection in response to diverse and open-ended natural language queries. However, manually annotating video segment timestamps and semantics is highly labor-intensive, making it challenging to obtain large-scale video-text data that meets our requirements. To address this, we adopt the following strategies:

\noindent(1) For TAD data, we integrate multiple relevant datasets during the pre-training phase to ensure comprehensive coverage of action categories and temporal annotations.

\noindent(2) For MR data, we leverage InternVid-10M-FLT \cite{wang2023internvid}, a dataset that employs a fully automated process for video segmentation and annotation, eliminating the need for manual intervention. However, due to inherent flaws in its automated annotation process, some annotations are of low quality. To mitigate this, we implement a filtering strategy: First, we discard video samples with UMT-SIM \cite{li2023unmasked} scores below $0.4$. Next, we curate a collection of over $1000$ keywords, ensuring they cover descriptions of most human activities in daily life. Finally, we filter the remaining video samples to include only those containing these keywords, resulting in a high-quality subset termed InternVid-MR. Table \ref{tab:datasets} provides detailed information about the datasets used in our pre-training phase.

\vspace{2mm}
\noindent\textbf{Loss Function.} Following \cite{zhang2022actionformer}, our model outputs segmentation results for each moment $t$ in the video, and we employ DIOU loss \cite{zheng2020distance} for distance regression to ensure precise temporal localization. Inspired by GLIP \cite{li2022grounded}, we utilize contrastive loss to align video moments with query representations for classification. Specifically, we dot product video features with query representations to predict logits for each query, followed by focal loss \cite{lin2017focal} for each logit. The final training loss is defined as:

\begin{equation}
\label{eq:loss}
    \mathcal{L}=\sum_t(\mathcal{L}_{cl}+{\lambda}\mathbb{P}_t\mathcal{L}_{reg})/T^+,
\end{equation}
where $\mathcal{L}_{cl}$ is the contrastive loss for classification; $\mathcal{L}_{reg}$ is the DIOU loss for distance regression; $\lambda$ is a coefficient balancing the classification and regression loss; $\mathbb{P}_t$ is an indicator function that denotes if a time step $t$ is a positive sample; $T^+$ is the total number of positive samples.

\begin{table}[t]
    \caption{A detailed list of Grounding-MD model variants. G-MD denotes Grounding-MD, while VMAE stands for VideoMAE.}
    \centering
    \scalebox{0.8}{
    \setlength\tabcolsep{3pt}
    \begin{tabular}{cccc}
        \toprule
        Model & Backbone & Pre-training Data & Res. \\
        \midrule
        G-MD-S (A) & VMAE-S & HACS & 160$^2$ \\
        G-MD-S (B) & VMAE-S & HACS, InternVid-MR-mini & 160$^2$ \\
        G-MD-L & VMAE-L & HACS, FineAction, InternVid-MR & 224$^2$ \\
        \bottomrule
    \end{tabular}
    }
\label{tab:variants}
\end{table}

\section{Experiments}
\label{sec:experiments}

\subsection{Benchmark Settings and Evaluation Metrics}

\noindent\textbf{Benchmark Settings.} We evaluated our framework extensively across three distinct settings: zero-shot domain transfer on ActivityNet \cite{caba2015activitynet}, open-vocabulary zero-shot on THUMOS14 \cite{idrees2017thumos}, and zero-shot domain transfer on ActivityNet-Captions \cite{krishna2017dense} and Charades-STA \cite{gao2017tall}.

\noindent\textbf{Evaluation Metrics.} For TAD, we report mAP at specific IoU thresholds, along with the average mAP, as the primary evaluation metrics. On ActivityNet, the IoU thresholds range from $0.5$ to $0.95$ in $10$ steps, while on THUMOS14, the thresholds are set to \{$0.3$,$0.4$,$0.5$,$0.6$,$0.7$\}. For MR, Recall@1 with IoU thresholds $0.5$ and $0.7$ are used on ActivityNet-Captions and Charades-STA.

\begin{table}[t]
    \caption{Zero-shot domain transfer and supervised fine-tuning on ActivityNet \cite{caba2015activitynet}, measured by mAP (\%) at different tIoU thresholds. E2E refers to end-to-end training. Specifically, AdaTAD$^*$ denotes that we pre-trained AdaTAD on HACS as the zero-shot baseline model. The best results are indicated in \textbf{bold}.}
    \centering
    \scalebox{0.8}{
    \setlength\tabcolsep{9pt}
    \begin{tabular}{ccccc}
        \toprule
        Method & Backbone & E2E & 0.5 & Avg. \\
        \midrule
        \multicolumn{5}{l}{\emph{Zero-Shot Domain Transfer}} \\
        \midrule
        AdaTAD$^*$ \cite{liu2024end} & VideoMAE-S & \checkmark & 45.5 & 27.3 \\
        G-MD-S (A) & VideoMAE-S & \checkmark & 46.6 & 28.9 \\
        G-MD-S (B) & VideoMAE-S & \checkmark & 48.3 & 29.7 \\
        G-MD-L & VideoMAE-L & \checkmark & \textbf{55.0} & \textbf{34.5} \\
        \midrule
        \midrule
        \multicolumn{5}{l}{\emph{Fine-Tuning}} \\
        \midrule
        TadTR \cite{liu2022end} & I3D & \texttimes & 49.1 & 32.3 \\
        BMN \cite{lin2019bmn} & TSN & \texttimes & 50.1 & 33.9 \\
        ActionFormer \cite{zhang2022actionformer} & SlowFast-R50 & \texttimes & 54.3 & 36.0 \\
        ASL \cite{shao2023action} & I3D & \texttimes & 54.1 & 36.2 \\
        TriDet \cite{shi2023tridet} & I3D & \texttimes & 54.7 & 36.8 \\
        InternVideo \cite{wang2022internvideo} & VideoMAE-H & \texttimes & - & 39.0 \\
        UniMD \cite{zeng2024unimd} & VideoMAE-H & \texttimes & 60.3 & 39.8 \\
        \midrule
        AFSD \cite{lin2021learning} & I3D & \checkmark & 52.4 & 34.4 \\
        E2E-TAD \cite{liu2022empirical} & SlowFast-R50 & \checkmark & 50.5 & 35.1 \\
        BasicTAD \cite{yang2023basictad} & SlowOnly-R50 & \checkmark & 51.2 & 33.1 \\
        TALLFormer \cite{cheng2022tallformer} & VideoSwin-B & \checkmark & 54.1 & 35.6 \\
        AdaTAD \cite{liu2024end} & VideoMAE-S & \checkmark & 56.2 & 37.8 \\
        AdaTAD \cite{liu2024end} & VideoMAEv2-g & \checkmark & 61.7 & 41.9 \\
        \midrule
        AdaTAD$^*$ \cite{liu2024end} & VideoMAE-S & \checkmark & 56.5 & 37.8 \\
        G-MD-S (A) & VideoMAE-S & \checkmark & 57.0 & 38.0 \\
        G-MD-S (B) & VideoMAE-S & \checkmark & 58.1 & 38.3 \\
        G-MD-L & VideoMAE-L & \checkmark & \textbf{62.0} & \textbf{42.1} \\
        \bottomrule
    \end{tabular}
    }
\label{tab:anet}
\end{table}

\subsection{Implementation Details}

We use VideoMAE \cite{tong2022videomae} with additional TIA \cite{liu2024end} as the video backbone and BERT \cite{devlin2019bert} as the text backbone, with the maximum number of text tokens set to $512$. We choose the Multi-scale Transformer Encoder from \cite{zhang2022actionformer} as the FPN module. The number of layers for CMFE, FPN, and TGFD are set to $3$, $5$, and $6$, respectively. The coefficient $\lambda$ in Equation \ref{eq:loss} is fixed at $1$.

We trained three model variants: Grounding-MD-S (A), which employs VideoMAE-S as the video backbone and is pre-trained on HACS; Grounding-MD-S (B), which also utilizes VideoMAE-S as the backbone but is pre-trained on both HACS and InternVid-MR-mini; and Grounding-MD-L, which adopts VideoMAE-L as the video backbone and is pre-trained on HACS, FineAction, and InternVid-MR. Here, InternVid-MR-mini represents a subset of InternVid-MR, consisting of 35000 videos randomly sampled from the full InternVid-MR dataset. The detailed configurations of these model variants are summarized in Table \ref{tab:variants}. All models are trained using AdamW optimizer \cite{loshchilov2017decoupled} for $10$ epochs, including $5$ epochs for warmup, on $4$ NVIDIA A100 GPUs with 40GB memory, implemented in PyTorch.

\noindent\textbf{Training Details.} Due to the maximum text token limit of $512$ in BERT \cite{devlin2019bert}, the number of action categories or event descriptions in a structured prompt cannot exceed this constraint. To address this, we restrict the number of action categories or event descriptions in a prompt to $35$ during training. For a temporal action detection video sample, we retain the positive classes present in the video and randomly select additional classes from the remaining categories as negative classes until the total number of classes reaches $35$. The order of these classes is then shuffled to generate a structured prompt. For moment retrieval video samples, since moment retrieval datasets lack predefined categories, we randomly select event descriptions from InternVid-MR as negative samples. The remaining operations align with those for temporal action detection samples.

\noindent\textbf{Evaluation Details.} Due to the maximum text token limit in BERT, we cannot include all $200$ predefined action categories of the ActivityNet \cite{caba2015activitynet} in a structured prompt during performance evaluation. To address this, we generate structured prompts in batches for the predefined action categories during the testing phase. Each batch is processed individually using BERT to generate the corresponding text tokens. These tokens are then concatenated to form the final token sequence.

\begin{table*}[htb]
    \caption{Open-vocabulary zero-shot performance on THUMOS14 \cite{idrees2017thumos}, measured by mAP (\%) at different tIoU thresholds. The \textcolor{gray}{gray} rows represent the open-vocabulary zero-shot results of Grounding-MD without pre-training. The best and the second-best results are indicated in \textbf{bold} and with \underline{underline}, respectively.}
    \centering
    \resizebox{1.0\textwidth}{!}{
    \begin{tabular}{ccccccccccccccc}
        \toprule
        \multirow{2}{*}{Method} & \multirow{2}{*}{Visual Backbone} & \multirow{2}{*}{Text Backbone} & \multicolumn{6}{c}{75\% Seen, 25\% Unseen} & \multicolumn{6}{c}{50\% Seen, 50\% Unseen} \\
        \cmidrule(lr){4-9}\cmidrule(lr){10-15}
        & & & 0.3 & 0.4 & 0.5 & 0.6 & 0.7 & Avg. & 0.3 & 0.4 & 0.5 & 0.6 & 0.7 & Avg. \\
        \midrule
        B-II \cite{radford2021learning} & CLIP-B & CLIP-B & 28.5 & 20.3 & 17.1 & 10.5 & 6.9 & 16.6 & 21.0 & 16.4 & 11.2 & 6.3 & 3.2 & 11.6 \\
        B-I \cite{radford2021learning} & CLIP-B & CLIP-B & 33.0 & 25.5 & 18.3 & 11.6 & 5.7 & 18.8 & 27.2 & 21.3 & 15.3 & 9.7 & 4.8 & 15.7 \\
        EffPrompt \cite{ju2022prompting} & CLIP-B & CLIP-B & 39.7 & 31.6 & 23.0 & 14.9 & 7.5 & 23.3 & 37.2 & 29.6 & 21.6 & 14.0 & 7.2 & 21.9 \\
        STALE \cite{nag2022zero} & CLIP-B & CLIP-B & \textbf{40.5} & 32.3 & 23.5 & 15.3 & 7.6 & 23.8 & \textbf{38.3} & 30.7 & 21.2 & 13.8 & 7.0 & 22.2 \\
        DeTAL \cite{li2024detal} & CLIP-B & CLIP-B & \underline{39.8} & \underline{33.6} & \underline{25.9} & \underline{17.4} & 9.9 & \underline{25.3} & \textbf{38.3} & \textbf{32.2} & \underline{24.4} & \underline{16.3} & \underline{9.0} & \underline{24.1} \\
        \midrule
        \textcolor{gray}{G-MD-S w/o pt} & \textcolor{gray}{VideoMAE-S} & \textcolor{gray}{BERT-B} & \textcolor{gray}{26.1} & \textcolor{gray}{23.4} & \textcolor{gray}{20.9} & \textcolor{gray}{16.8} & \textcolor{gray}{\underline{11.5}} & \textcolor{gray}{19.7} & \textcolor{gray}{15.4} & \textcolor{gray}{13.3} & \textcolor{gray}{11.8} & \textcolor{gray}{10.3} & \textcolor{gray}{8.6} & \textcolor{gray}{11.9} \\
        G-MD-S (B) w/ pt & VideoMAE-S & BERT-B & 39.6 & \textbf{33.7} & \textbf{27.3} & \textbf{20.3} & \textbf{14.5} & \textbf{27.1} & \underline{37.9} & \underline{32.0} & \textbf{26.3} & \textbf{19.7} & \textbf{13.9} & \textbf{26.0} \\
        \bottomrule
    \end{tabular}
    }
\label{tab:zsthumos}
\end{table*}

\subsection{Zero-Shot Domain Transfer on ActivityNet}

We conducted comprehensive experiments on ActivityNet \cite{caba2015activitynet} to evaluate the transfer capabilities of models across common categories. The evaluation was performed in two settings: zero-shot domain transfer and supervised fine-tuning. Given that HACS \cite{zhao2019hacs} encompasses all categories present in ActivityNet \cite{caba2015activitynet}, we used AdaTAD \cite{liu2024end} pre-trained on HACS as a zero-shot baseline for comparison. The results, summarized in Table \ref{tab:anet}, demonstrate that Grounding-MD models achieve strong performance in both zero-shot and supervised settings.

In the zero-shot domain transfer setting, when HACS was employed as the pre-training dataset, Grounding-MD-S achieved a $1.6\%$ higher mAP than AdaTAD. Furthermore, integrating the InternVid-MR-mini dataset into the pre-training process enhanced the mAP by an additional $0.8\%$, demonstrating the benefits of incorporating diverse video-text data for improved generalization. In the supervised fine-tuning setting, Grounding-MD-L, which utilized VideoMAE-L as the video backbone, outperformed AdaTAD (which employed VideoMAEv2-g as its video backbone) by $0.2\%$ in terms of mAP.

\subsection{Open-Vocabulary Zero-Shot on THUMOS14}

Our evaluation on THUMOS14 \cite{idrees2017thumos} is conducted under two settings: 1) open-vocabulary zero-shot, which assesses the ability to generalize to unseen action categories, and 2) supervised fine-tuning.

For the open-vocabulary zero-shot setting, we follow the standard proposed in \cite{ju2022prompting} by dividing THUMOS14 into two subsets based on action categories. We randomly sampled the dataset $10$ times for two scenarios (75\% seen and 25\% unseen, 50\% seen and 50\% unseen, respectively). The average results across these samplings are reported as the final metrics. The open-vocabulary zero-shot performance on THUMOS14 is shown in Table \ref{tab:zsthumos}. Unlike other approaches that utilize CLIP \cite{radford2021learning} as the backbone for both visual and textual data—which inherently establishes a semantic connection between vision and language—the combination of VideoMAE and BERT lacks such an intrinsic link. As a result, Grounding-MD initially exhibits poor open-vocabulary zero-shot performance without pre-training. However, after pre-training, it achieves superior open-vocabulary zero-shot performance, clearly demonstrating its effectiveness in video-text alignment. The supervised fine-tuning results are shown in Table \ref{tab:ftthumos}, Grounding-MD-S achieves a significant performance improvement compared to other methods. Notably, on the most challenging IoU threshold of $0.7$ in the mAP metric, Grounding-MD-S outperforms TriDet \cite{shi2023tridet} by $3.3$ \% points and surpasses the state-of-the-art method AdaTAD by $3.8$ \% points, with the same video backbone (VideoMAE-S).

\begin{table}[t]
    \caption{Supervised fine-tuning on THUMOS14 \cite{idrees2017thumos}, measured by mAP (\%) at different tIoU thresholds. The best and the second-best results are indicated in \textbf{bold} and with \underline{underline}, respectively.}
    \centering
    \setlength\tabcolsep{3pt}
    \resizebox{0.47\textwidth}{!}{
    \begin{tabular}{ccccccccc}
        \toprule
        Method & Backbone & E2E & 0.3 & 0.4 & 0.5 & 0.6 & 0.7 & Avg. \\
        \midrule
        BMN \cite{lin2019bmn} & TSN & \texttimes & 56.0 & 47.4 & 38.8 & 29.7 & 20.5 & 38.5 \\
        TadTR \cite{liu2022end} & I3D & \texttimes & 62.4 & 57.4 & 49.2 & 37.8 & 26.3 & 46.6 \\
        ActionFormer \cite{zhang2022actionformer} & I3D & \texttimes & 82.1 & 77.8 & 71.0 & 59.4 & 43.9 & 66.8 \\
        ASL \cite{shao2023action} & I3D & \texttimes & 83.1 & 79.0 & 71.7 & 59.7 & 45.8 & 67.9 \\
        TriDet \cite{shi2023tridet} & I3D & \texttimes & 83.6 & 80.1 & \underline{72.9} & \underline{62.4} & \underline{47.4} & \underline{69.3} \\
        InternVideo \cite{wang2022internvideo} & VideoMAE-H & \texttimes & - & - & - & - & - & 71.5 \\
        \midrule
        AFSD \cite{lin2021learning} & I3D & \checkmark & 67.3 & 62.4 & 55.5 & 43.7 & 31.1 & 52.0 \\
        E2E-TAD \cite{liu2022empirical} & SlowFast-R50 & \checkmark & 69.4 & 64.3 & 56.0 & 46.4 & 34.9 & 54.2 \\
        BasicTAD \cite{yang2023basictad} & SlowOnly-R50 & \checkmark & 75.5 & 70.8 & 63.5 & 50.9 & 37.4 & 59.6 \\
        TALLFormer \cite{cheng2022tallformer} & VideoSwin-B & \checkmark & 76.0 & - & 63.2 & - & 34.5 & 59.2 \\
        AdaTAD \cite{liu2024end} & VideoMAE-S & \checkmark & \underline{84.5} & \underline{80.2} & 71.6 & 60.9 & 46.9 & 68.8 \\
        \midrule
        G-MD-S (B) & VideoMAE-S & \checkmark & \textbf{84.8} & \textbf{80.3} & \textbf{73.1} & \textbf{62.4} & \textbf{50.5} & \textbf{70.2} \\
        \bottomrule
    \end{tabular}
    }
\label{tab:ftthumos}
\end{table}

\begin{table*}[t]
    \caption{Zero-shot domain transfer on ActivityNet-Captions \cite{krishna2017dense} and Charades-STA \cite{gao2017tall}, measured by Recall@1 (\%) with tIoU thresholds $0.5$ and $0.7$. The best and the second-best results are indicated in \textbf{bold} and with \underline{underline}, respectively.}
    \centering
    \scalebox{0.8}{
    \begin{tabular}{ccccccc}
        \toprule
        \multirow{2}{*}{Method} & \multirow{2}{*}{Param} & \multirow{2}{*}{Pre-training Data} & \multicolumn{2}{c}{ANet-Captions} & \multicolumn{2}{c}{Charades-STA} \\
        \cmidrule(lr){4-5}\cmidrule(lr){6-7}
        & & & R@0.5 & R@0.7 & R@0.5 & R@0.7 \\
        \midrule
        Valley \cite{luo2023valley} & 7B & WebVid2M \cite{bain2021frozen}, CC3M \cite{sharma2018conceptual} & 13.7 & 8.1 & 1.8 & 0.3 \\
        Video-LLaMA \cite{zhang2023video} & 7B & WebVid2M \cite{bain2021frozen}, CC3M \cite{sharma2018conceptual} & 10.8 & 4.9 & 10.6 & 3.4 \\
        VideoChat \cite{li2023videochat} & 7B & WebVid10M \cite{bain2021frozen}, CC12M \cite{changpinyo2021conceptual} & 12.6 & 6.0 & 8.6 & 0.0 \\
        VideoChat2 \cite{li2024mvbench} & 7B & WebVid10M \cite{bain2021frozen}, InternVid-10M \cite{wang2023internvid}, CC12M \cite{changpinyo2021conceptual} & \textbf{27.8} & 9.3 & 14.3 & 3.8 \\
        Momentor \cite{qian2024momentor} & 7B & Moment-10M \cite{qian2024momentor} & 23.0 & \underline{12.4} & \textbf{26.6} & \underline{11.6} \\
        \midrule
        G-MD-L & 0.5B & HACS, FineAction, InternVid-MR & \underline{25.3} & \textbf{12.8} & \underline{24.5} & \textbf{13.0} \\
        \bottomrule
    \end{tabular}
    }
\label{tab:zsmr}
\end{table*}

\begin{table}[t]
    \caption{Comparative results of supervised fine-tuning on ActivityNet-Captions \cite{krishna2017dense} and Charades-STA \cite{gao2017tall} against other end-to-end moment retrieval methods, measured by Recall@1 (\%) with tIoU thresholds $0.5$ and $0.7$. The best and the second-best results are indicated in \textbf{bold} and with \underline{underline}, respectively.}
    \centering
    \setlength\tabcolsep{10pt}
    \scalebox{0.8}{
    \begin{tabular}{ccccc}
        \toprule
        \multirow{2}{*}{Method} & \multicolumn{2}{c}{ANet-Captions} & \multicolumn{2}{c}{Charades-STA} \\
        \cmidrule(lr){2-3}\cmidrule(lr){4-5}
        & R@0.5 & R@0.7 & R@0.5 & R@0.7 \\
        \midrule
        VDI \cite{luo2023towards} & 48.1 & 28.8 & 52.3 & 31.4 \\
        UnLoc-L \cite{yan2023unloc} & \underline{48.3} & \underline{30.2} & \underline{60.8} & 38.4 \\
        UniVTG \cite{lin2023univtg} & - & - & 60.2 & \underline{38.6} \\
        \midrule
        G-MD-L & \textbf{49.8} & \textbf{31.4} & \textbf{62.0} & \textbf{39.9} \\
        \bottomrule
    \end{tabular}
    }
\label{tab:ftmr}
\end{table}

\subsection{Zero-Shot Domain Transfer on ActivityNet-Captions and Charades-STA}

We evaluated the zero-shot domain transfer of Grounding-MD in the MD task, as shown in Table \ref{tab:zsmr}. Despite having a significantly smaller model size and less training data, Grounding-MD achieved superior performance compared to Video-LLM-based methods. Particularly on the more challenging Recall@0.7 metric, Grounding-MD outperforms Momentor \cite{qian2024momentor} by $1.4\%$ points on the Charades-STA dataset and by $0.3\%$ points on the ActivityNet-Captions dataset. As illustrated in Table \ref{tab:ftmr}, the supervised fine-tuning of Grounding-MD demonstrates substantial performance improvements. Compared to previous end-to-end moment retrieval approaches, Grounding-MD achieves state-of-the-art results across all evaluation metrics.

\begin{table}[htb]
    \caption{Ablation experiments conducted on the ActivityNet \cite{caba2015activitynet} dataset. All variant models were pre-trained on HACS and InternVid-MR-mini using VideoMAE-S as the video backbone.}
    \centering
    \setlength\tabcolsep{16pt}
    \scalebox{0.8}{
    \begin{tabular}{ccccccc}
        \toprule
        Model & 0.5 & Avg. \\
        \midrule
        Grounding-MD (w/ Average Pooling) & 48.3 & 29.7 \\
        w/o CMFE & 45.9 & 27.6 \\
        w/o TGFD & 47.1 & 28.7 \\
        \midrule
        w/ Attentive Pooling & 45.2 & 26.8 \\
        w/ Max Pooling & 47.4 & 29.1 \\
        \bottomrule
    \end{tabular}
    }
\label{tab:abl}
\end{table}

\subsection{Ablation Study}

We conduct ablation studies to validate the effectiveness of the components in Grounding-MD, which is based on deep early-late video-text fusion. To achieve this, we systematically remove or replace specific modules in different variants of our model and analyze their impact on performance.

As shown in rows $2$ and $3$ of Table \ref{tab:abl}, the absence of either CMFE or TGFD leads to a significant drop in model performance, with CMFE having a particularly pronounced impact. CMFE is critical because the video and text backbone are inherently independent, with no semantic connection between the extracted video and text features. CMFE establishes the first channel of interaction between these modalities, enabling early fusion. Through this process, video features become more focused on entities related to semantic information, while text features emphasize tokens containing the main semantic content, thereby laying the foundation for effective cross-modal alignment.

By comparing rows $1$, $4$, and $5$ of Table \ref{tab:abl}, we observe that average pooling achieved the best performance. The ineffectiveness of attention pooling can be attributed to the weighted averaging it introduces, which may disrupt the text features enhanced by CMFE. This disruption could shift the focus of the text features away from the key semantic information initially aligned during early fusion. In contrast, average pooling, despite its simplicity, preserves the integrity of feature alignment established by CMFE, maintaining semantic coherence between video and text modalities. Thus, average pooling proves to be more effective for our task, ensuring stable and accurate alignment between video and textual information. This finding highlights the importance of preserving the semantic structure established during early fusion for robust moment detection.

\subsection{Limitations}

Although Grounding-MD demonstrates remarkable performance as a unified open-world moment detection method, it still faces specific challenges and limitations that need to be addressed. One potential constraint lies in scaling Grounding-MD by incorporating larger video backbones and leveraging more extensive datasets to enhance its open-world moment detection capabilities and applicability. However, the pre-training phase demands substantial computational resources, which inevitably poses a barrier to scalability. To overcome this, our future work will integrate masked video modeling to develop an efficient video-language pre-training strategy. This approach aims to reduce computational resource dependency, thereby enabling the expansion of the model's video backbone and the scale of pre-training data.

\section{Conclusion}
\label{sec:conclusion}

In this paper, we introduced Grounding-MD, a grounded video-language pre-training framework for open-world moment detection. By integrating temporal action detection and moment retrieval into a single model, Grounding-MD addresses the limitations of existing approaches that are confined to closed-set settings. Our key innovations include a structured prompt design that accommodates an arbitrary number of open-ended natural language queries and a cross-modality fusion mechanism that ensures deep alignment between video and text modalities. These advancements enable Grounding-MD to achieve superior performance in both zero-shot and supervised settings, as demonstrated by extensive experiments on multiple benchmarks. Grounding-MD represents a significant step forward in video understanding, bridging the gap between temporal action detection and moment retrieval while addressing the challenges of open-world scenarios. 
{
    \small
    \bibliographystyle{ieeenat_fullname}
    \bibliography{main}

\begin{thebibliography}{60}
\providecommand{\natexlab}[1]{#1}
\providecommand{\url}[1]{\texttt{#1}}
\expandafter\ifx\csname urlstyle\endcsname\relax
  \providecommand{\doi}[1]{doi: #1}\else
  \providecommand{\doi}{doi: \begingroup \urlstyle{rm}\Url}\fi

\bibitem[Anne~Hendricks et~al.(2017)Anne~Hendricks, Wang, Shechtman, Sivic,
  Darrell, and Russell]{anne2017localizing}
Lisa Anne~Hendricks, Oliver Wang, Eli Shechtman, Josef Sivic, Trevor Darrell,
  and Bryan Russell.
\newblock Localizing moments in video with natural language.
\newblock In \emph{Proceedings of the IEEE international conference on computer
  vision}, pages 5803--5812, 2017.

\bibitem[Bai et~al.(2020)Bai, Wang, Tong, Yang, Liu, and Liu]{bai2020boundary}
Yueran Bai, Yingying Wang, Yunhai Tong, Yang Yang, Qiyue Liu, and Junhui Liu.
\newblock Boundary content graph neural network for temporal action proposal
  generation.
\newblock In \emph{Computer Vision--ECCV 2020: 16th European Conference,
  Glasgow, UK, August 23--28, 2020, Proceedings, Part XXVIII 16}, pages
  121--137. Springer, 2020.

\bibitem[Bain et~al.(2021)Bain, Nagrani, Varol, and Zisserman]{bain2021frozen}
Max Bain, Arsha Nagrani, G{\"u}l Varol, and Andrew Zisserman.
\newblock Frozen in time: A joint video and image encoder for end-to-end
  retrieval.
\newblock In \emph{Proceedings of the IEEE/CVF international conference on
  computer vision}, pages 1728--1738, 2021.

\bibitem[Caba~Heilbron et~al.(2015)Caba~Heilbron, Escorcia, Ghanem, and
  Carlos~Niebles]{caba2015activitynet}
Fabian Caba~Heilbron, Victor Escorcia, Bernard Ghanem, and Juan Carlos~Niebles.
\newblock Activitynet: A large-scale video benchmark for human activity
  understanding.
\newblock In \emph{Proceedings of the ieee conference on computer vision and
  pattern recognition}, pages 961--970, 2015.

\bibitem[Changpinyo et~al.(2021)Changpinyo, Sharma, Ding, and
  Soricut]{changpinyo2021conceptual}
Soravit Changpinyo, Piyush Sharma, Nan Ding, and Radu Soricut.
\newblock Conceptual 12m: Pushing web-scale image-text pre-training to
  recognize long-tail visual concepts.
\newblock In \emph{Proceedings of the IEEE/CVF conference on computer vision
  and pattern recognition}, pages 3558--3568, 2021.

\bibitem[Cheng and Bertasius(2022)]{cheng2022tallformer}
Feng Cheng and Gedas Bertasius.
\newblock Tallformer: Temporal action localization with a long-memory
  transformer.
\newblock In \emph{European Conference on Computer Vision}, pages 503--521.
  Springer, 2022.

\bibitem[Cheng et~al.(2023)Cheng, Wang, Lei, Crandall, Bansal, and
  Bertasius]{cheng2023vindlu}
Feng Cheng, Xizi Wang, Jie Lei, David Crandall, Mohit Bansal, and Gedas
  Bertasius.
\newblock Vindlu: A recipe for effective video-and-language pretraining.
\newblock In \emph{Proceedings of the IEEE/CVF Conference on Computer Vision
  and Pattern Recognition}, pages 10739--10750, 2023.

\bibitem[Devlin et~al.(2019)Devlin, Chang, Lee, and Toutanova]{devlin2019bert}
Jacob Devlin, Ming-Wei Chang, Kenton Lee, and Kristina Toutanova.
\newblock Bert: Pre-training of deep bidirectional transformers for language
  understanding.
\newblock In \emph{Proceedings of the 2019 conference of the North American
  chapter of the association for computational linguistics: human language
  technologies, volume 1 (long and short papers)}, pages 4171--4186, 2019.

\bibitem[Fan et~al.(2018)Fan, Huang, Gan, Ermon, Gong, and Huang]{fan2018end}
Lijie Fan, Wenbing Huang, Chuang Gan, Stefano Ermon, Boqing Gong, and Junzhou
  Huang.
\newblock End-to-end learning of motion representation for video understanding.
\newblock In \emph{Proceedings of the IEEE conference on computer vision and
  pattern recognition}, pages 6016--6025, 2018.

\bibitem[Gao et~al.(2017)Gao, Sun, Yang, and Nevatia]{gao2017tall}
Jiyang Gao, Chen Sun, Zhenheng Yang, and Ram Nevatia.
\newblock Tall: Temporal activity localization via language query.
\newblock In \emph{Proceedings of the IEEE international conference on computer
  vision}, pages 5267--5275, 2017.

\bibitem[Idrees et~al.(2017)Idrees, Zamir, Jiang, Gorban, Laptev, Sukthankar,
  and Shah]{idrees2017thumos}
Haroon Idrees, Amir~R Zamir, Yu-Gang Jiang, Alex Gorban, Ivan Laptev, Rahul
  Sukthankar, and Mubarak Shah.
\newblock The thumos challenge on action recognition for videos “in the
  wild”.
\newblock \emph{Computer Vision and Image Understanding}, 155:\penalty0 1--23,
  2017.

\bibitem[Jiang et~al.(2024)Jiang, Li, Zeng, Ren, Liu, and Zhang]{jiang2024t}
Qing Jiang, Feng Li, Zhaoyang Zeng, Tianhe Ren, Shilong Liu, and Lei Zhang.
\newblock T-rex2: Towards generic object detection via text-visual prompt
  synergy.
\newblock In \emph{European Conference on Computer Vision}, pages 38--57.
  Springer, 2024.

\bibitem[Ju et~al.(2022)Ju, Han, Zheng, Zhang, and Xie]{ju2022prompting}
Chen Ju, Tengda Han, Kunhao Zheng, Ya Zhang, and Weidi Xie.
\newblock Prompting visual-language models for efficient video understanding.
\newblock In \emph{European Conference on Computer Vision}, pages 105--124.
  Springer, 2022.

\bibitem[Krishna et~al.(2017)Krishna, Hata, Ren, Fei-Fei, and
  Carlos~Niebles]{krishna2017dense}
Ranjay Krishna, Kenji Hata, Frederic Ren, Li Fei-Fei, and Juan Carlos~Niebles.
\newblock Dense-captioning events in videos.
\newblock In \emph{Proceedings of the IEEE international conference on computer
  vision}, pages 706--715, 2017.

\bibitem[Li et~al.(2023{\natexlab{a}})Li, He, Wang, Li, Wang, Luo, Wang, Wang,
  and Qiao]{li2023videochat}
KunChang Li, Yinan He, Yi Wang, Yizhuo Li, Wenhai Wang, Ping Luo, Yali Wang,
  Limin Wang, and Yu Qiao.
\newblock Videochat: Chat-centric video understanding.
\newblock \emph{arXiv preprint arXiv:2305.06355}, 2023{\natexlab{a}}.

\bibitem[Li et~al.(2023{\natexlab{b}})Li, Wang, Li, Wang, He, Wang, and
  Qiao]{li2023unmasked}
Kunchang Li, Yali Wang, Yizhuo Li, Yi Wang, Yinan He, Limin Wang, and Yu Qiao.
\newblock Unmasked teacher: Towards training-efficient video foundation models.
\newblock In \emph{Proceedings of the IEEE/CVF International Conference on
  Computer Vision}, pages 19948--19960, 2023{\natexlab{b}}.

\bibitem[Li et~al.(2024{\natexlab{a}})Li, Wang, He, Li, Wang, Liu, Wang, Xu,
  Chen, Luo, et~al.]{li2024mvbench}
Kunchang Li, Yali Wang, Yinan He, Yizhuo Li, Yi Wang, Yi Liu, Zun Wang, Jilan
  Xu, Guo Chen, Ping Luo, et~al.
\newblock Mvbench: A comprehensive multi-modal video understanding benchmark.
\newblock In \emph{Proceedings of the IEEE/CVF Conference on Computer Vision
  and Pattern Recognition}, pages 22195--22206, 2024{\natexlab{a}}.

\bibitem[Li et~al.(2022)Li, Zhang, Zhang, Yang, Li, Zhong, Wang, Yuan, Zhang,
  Hwang, et~al.]{li2022grounded}
Liunian~Harold Li, Pengchuan Zhang, Haotian Zhang, Jianwei Yang, Chunyuan Li,
  Yiwu Zhong, Lijuan Wang, Lu Yuan, Lei Zhang, Jenq-Neng Hwang, et~al.
\newblock Grounded language-image pre-training.
\newblock In \emph{Proceedings of the IEEE/CVF conference on computer vision
  and pattern recognition}, pages 10965--10975, 2022.

\bibitem[Li et~al.(2024{\natexlab{b}})Li, Zhong, Song, Li, Ma, and
  Zhang]{li2024detal}
Zhiheng Li, Yujie Zhong, Ran Song, Tianjiao Li, Lin Ma, and Wei Zhang.
\newblock Detal: open-vocabulary temporal action localization with decoupled
  networks.
\newblock \emph{IEEE Transactions on Pattern Analysis and Machine
  Intelligence}, 2024{\natexlab{b}}.

\bibitem[Lin et~al.(2021)Lin, Xu, Luo, Wang, Tai, Wang, Li, Huang, and
  Fu]{lin2021learning}
Chuming Lin, Chengming Xu, Donghao Luo, Yabiao Wang, Ying Tai, Chengjie Wang,
  Jilin Li, Feiyue Huang, and Yanwei Fu.
\newblock Learning salient boundary feature for anchor-free temporal action
  localization.
\newblock In \emph{Proceedings of the IEEE/CVF conference on computer vision
  and pattern recognition}, pages 3320--3329, 2021.

\bibitem[Lin et~al.(2019{\natexlab{a}})Lin, Gan, and Han]{lin2019tsm}
Ji Lin, Chuang Gan, and Song Han.
\newblock Tsm: Temporal shift module for efficient video understanding.
\newblock In \emph{Proceedings of the IEEE/CVF international conference on
  computer vision}, pages 7083--7093, 2019{\natexlab{a}}.

\bibitem[Lin et~al.(2023)Lin, Zhang, Chen, Pramanick, Gao, Wang, Yan, and
  Shou]{lin2023univtg}
Kevin~Qinghong Lin, Pengchuan Zhang, Joya Chen, Shraman Pramanick, Difei Gao,
  Alex~Jinpeng Wang, Rui Yan, and Mike~Zheng Shou.
\newblock Univtg: Towards unified video-language temporal grounding.
\newblock In \emph{Proceedings of the IEEE/CVF International Conference on
  Computer Vision}, pages 2794--2804, 2023.

\bibitem[Lin et~al.(2017{\natexlab{a}})Lin, Zhao, and Shou]{lin2017single}
Tianwei Lin, Xu Zhao, and Zheng Shou.
\newblock Single shot temporal action detection.
\newblock In \emph{Proceedings of the 25th ACM international conference on
  Multimedia}, pages 988--996, 2017{\natexlab{a}}.

\bibitem[Lin et~al.(2019{\natexlab{b}})Lin, Liu, Li, Ding, and Wen]{lin2019bmn}
Tianwei Lin, Xiao Liu, Xin Li, Errui Ding, and Shilei Wen.
\newblock Bmn: Boundary-matching network for temporal action proposal
  generation.
\newblock In \emph{Proceedings of the IEEE/CVF international conference on
  computer vision}, pages 3889--3898, 2019{\natexlab{b}}.

\bibitem[Lin et~al.(2017{\natexlab{b}})Lin, Goyal, Girshick, He, and
  Doll{\'a}r]{lin2017focal}
Tsung-Yi Lin, Priya Goyal, Ross Girshick, Kaiming He, and Piotr Doll{\'a}r.
\newblock Focal loss for dense object detection.
\newblock In \emph{Proceedings of the IEEE international conference on computer
  vision}, pages 2980--2988, 2017{\natexlab{b}}.

\bibitem[Liu et~al.(2024{\natexlab{a}})Liu, Zeng, Ren, Li, Zhang, Yang, Jiang,
  Li, Yang, Su, et~al.]{liu2024grounding}
Shilong Liu, Zhaoyang Zeng, Tianhe Ren, Feng Li, Hao Zhang, Jie Yang, Qing
  Jiang, Chunyuan Li, Jianwei Yang, Hang Su, et~al.
\newblock Grounding dino: Marrying dino with grounded pre-training for open-set
  object detection.
\newblock In \emph{European Conference on Computer Vision}, pages 38--55.
  Springer, 2024{\natexlab{a}}.

\bibitem[Liu et~al.(2024{\natexlab{b}})Liu, Zhang, Zhao, and
  Ghanem]{liu2024end}
Shuming Liu, Chen-Lin Zhang, Chen Zhao, and Bernard Ghanem.
\newblock End-to-end temporal action detection with 1b parameters across 1000
  frames.
\newblock In \emph{Proceedings of the IEEE/CVF conference on computer vision
  and pattern recognition}, pages 18591--18601, 2024{\natexlab{b}}.

\bibitem[Liu et~al.(2022{\natexlab{a}})Liu, Bai, and Bai]{liu2022empirical}
Xiaolong Liu, Song Bai, and Xiang Bai.
\newblock An empirical study of end-to-end temporal action detection.
\newblock In \emph{Proceedings of the IEEE/CVF Conference on Computer Vision
  and Pattern Recognition}, pages 20010--20019, 2022{\natexlab{a}}.

\bibitem[Liu et~al.(2022{\natexlab{b}})Liu, Wang, Hu, Tang, Zhang, Bai, and
  Bai]{liu2022end}
Xiaolong Liu, Qimeng Wang, Yao Hu, Xu Tang, Shiwei Zhang, Song Bai, and Xiang
  Bai.
\newblock End-to-end temporal action detection with transformer.
\newblock \emph{IEEE Transactions on Image Processing}, 31:\penalty0
  5427--5441, 2022{\natexlab{b}}.

\bibitem[Liu et~al.(2022{\natexlab{c}})Liu, Li, Wu, Chen, Shan, and
  Qie]{liu2022umt}
Ye Liu, Siyuan Li, Yang Wu, Chang-Wen Chen, Ying Shan, and Xiaohu Qie.
\newblock Umt: Unified multi-modal transformers for joint video moment
  retrieval and highlight detection.
\newblock In \emph{Proceedings of the IEEE/CVF Conference on Computer Vision
  and Pattern Recognition}, pages 3042--3051, 2022{\natexlab{c}}.

\bibitem[Liu et~al.(2022{\natexlab{d}})Liu, Wang, Wang, Ma, and
  Qiao]{liu2022fineaction}
Yi Liu, Limin Wang, Yali Wang, Xiao Ma, and Yu Qiao.
\newblock Fineaction: A fine-grained video dataset for temporal action
  localization.
\newblock \emph{IEEE transactions on image processing}, 31:\penalty0
  6937--6950, 2022{\natexlab{d}}.

\bibitem[Loshchilov and Hutter(2017)]{loshchilov2017decoupled}
Ilya Loshchilov and Frank Hutter.
\newblock Decoupled weight decay regularization.
\newblock \emph{arXiv preprint arXiv:1711.05101}, 2017.

\bibitem[Luo et~al.(2023{\natexlab{a}})Luo, Huang, Gong, Jin, and
  Liu]{luo2023towards}
Dezhao Luo, Jiabo Huang, Shaogang Gong, Hailin Jin, and Yang Liu.
\newblock Towards generalisable video moment retrieval: Visual-dynamic
  injection to image-text pre-training.
\newblock In \emph{Proceedings of the IEEE/CVF Conference on Computer Vision
  and Pattern Recognition}, pages 23045--23055, 2023{\natexlab{a}}.

\bibitem[Luo et~al.(2023{\natexlab{b}})Luo, Zhao, Yang, Dong, Li, Lu, Wang, Hu,
  Qiu, and Wei]{luo2023valley}
Ruipu Luo, Ziwang Zhao, Min Yang, Junwei Dong, Da Li, Pengcheng Lu, Tao Wang,
  Linmei Hu, Minghui Qiu, and Zhongyu Wei.
\newblock Valley: Video assistant with large language model enhanced ability.
\newblock \emph{arXiv preprint arXiv:2306.07207}, 2023{\natexlab{b}}.

\bibitem[Mun et~al.(2020)Mun, Cho, and Han]{mun2020local}
Jonghwan Mun, Minsu Cho, and Bohyung Han.
\newblock Local-global video-text interactions for temporal grounding.
\newblock In \emph{Proceedings of the IEEE/CVF Conference on Computer Vision
  and Pattern Recognition}, pages 10810--10819, 2020.

\bibitem[Nag et~al.(2022)Nag, Zhu, Song, and Xiang]{nag2022zero}
Sauradip Nag, Xiatian Zhu, Yi-Zhe Song, and Tao Xiang.
\newblock Zero-shot temporal action detection via vision-language prompting.
\newblock In \emph{European conference on computer vision}, pages 681--697.
  Springer, 2022.

\bibitem[Qian et~al.(2024)Qian, Li, Wu, Ye, Fei, Chua, Zhuang, and
  Tang]{qian2024momentor}
Long Qian, Juncheng Li, Yu Wu, Yaobo Ye, Hao Fei, Tat-Seng Chua, Yueting
  Zhuang, and Siliang Tang.
\newblock Momentor: Advancing video large language model with fine-grained
  temporal reasoning.
\newblock \emph{arXiv preprint arXiv:2402.11435}, 2024.

\bibitem[Radford et~al.(2021)Radford, Kim, Hallacy, Ramesh, Goh, Agarwal,
  Sastry, Askell, Mishkin, Clark, et~al.]{radford2021learning}
Alec Radford, Jong~Wook Kim, Chris Hallacy, Aditya Ramesh, Gabriel Goh,
  Sandhini Agarwal, Girish Sastry, Amanda Askell, Pamela Mishkin, Jack Clark,
  et~al.
\newblock Learning transferable visual models from natural language
  supervision.
\newblock In \emph{International conference on machine learning}, pages
  8748--8763. PmLR, 2021.

\bibitem[Shao et~al.(2023)Shao, Wang, Quan, Zheng, Yang, and
  Yang]{shao2023action}
Jiayi Shao, Xiaohan Wang, Ruijie Quan, Junjun Zheng, Jiang Yang, and Yi Yang.
\newblock Action sensitivity learning for temporal action localization.
\newblock In \emph{Proceedings of the IEEE/CVF International Conference on
  Computer Vision}, pages 13457--13469, 2023.

\bibitem[Sharma et~al.(2018)Sharma, Ding, Goodman, and
  Soricut]{sharma2018conceptual}
Piyush Sharma, Nan Ding, Sebastian Goodman, and Radu Soricut.
\newblock Conceptual captions: A cleaned, hypernymed, image alt-text dataset
  for automatic image captioning.
\newblock In \emph{Proceedings of the 56th Annual Meeting of the Association
  for Computational Linguistics (Volume 1: Long Papers)}, pages 2556--2565,
  2018.

\bibitem[Shi et~al.(2023)Shi, Zhong, Cao, Ma, Li, and Tao]{shi2023tridet}
Dingfeng Shi, Yujie Zhong, Qiong Cao, Lin Ma, Jia Li, and Dacheng Tao.
\newblock Tridet: Temporal action detection with relative boundary modeling.
\newblock In \emph{Proceedings of the IEEE/CVF Conference on Computer Vision
  and Pattern Recognition}, pages 18857--18866, 2023.

\bibitem[Tong et~al.(2022)Tong, Song, Wang, and Wang]{tong2022videomae}
Zhan Tong, Yibing Song, Jue Wang, and Limin Wang.
\newblock Videomae: Masked autoencoders are data-efficient learners for
  self-supervised video pre-training.
\newblock \emph{Advances in neural information processing systems},
  35:\penalty0 10078--10093, 2022.

\bibitem[Vaswani et~al.(2017)Vaswani, Shazeer, Parmar, Uszkoreit, Jones, Gomez,
  Kaiser, and Polosukhin]{vaswani2017attention}
Ashish Vaswani, Noam Shazeer, Niki Parmar, Jakob Uszkoreit, Llion Jones,
  Aidan~N Gomez, {\L}ukasz Kaiser, and Illia Polosukhin.
\newblock Attention is all you need.
\newblock \emph{Advances in neural information processing systems}, 30, 2017.

\bibitem[Wang et~al.(2021)Wang, Xing, and Liu]{wang2021actionclip}
Mengmeng Wang, Jiazheng Xing, and Yong Liu.
\newblock Actionclip: A new paradigm for video action recognition.
\newblock \emph{arXiv preprint arXiv:2109.08472}, 2021.

\bibitem[Wang et~al.(2022)Wang, Li, Li, He, Huang, Zhao, Zhang, Xu, Liu, Wang,
  et~al.]{wang2022internvideo}
Yi Wang, Kunchang Li, Yizhuo Li, Yinan He, Bingkun Huang, Zhiyu Zhao, Hongjie
  Zhang, Jilan Xu, Yi Liu, Zun Wang, et~al.
\newblock Internvideo: General video foundation models via generative and
  discriminative learning.
\newblock \emph{arXiv preprint arXiv:2212.03191}, 2022.

\bibitem[Wang et~al.(2023)Wang, He, Li, Li, Yu, Ma, Li, Chen, Chen, Wang,
  et~al.]{wang2023internvid}
Yi Wang, Yinan He, Yizhuo Li, Kunchang Li, Jiashuo Yu, Xin Ma, Xinhao Li, Guo
  Chen, Xinyuan Chen, Yaohui Wang, et~al.
\newblock Internvid: A large-scale video-text dataset for multimodal
  understanding and generation.
\newblock \emph{arXiv preprint arXiv:2307.06942}, 2023.

\bibitem[Xia et~al.(2022)Xia, Wang, Zhou, Zheng, and Tang]{xia2022learning}
Kun Xia, Le Wang, Sanping Zhou, Nanning Zheng, and Wei Tang.
\newblock Learning to refactor action and co-occurrence features for temporal
  action localization.
\newblock In \emph{Proceedings of the IEEE/CVF conference on computer vision
  and pattern recognition}, pages 13884--13893, 2022.

\bibitem[Yan et~al.(2023)Yan, Xiong, Nagrani, Arnab, Wang, Ge, Ross, and
  Schmid]{yan2023unloc}
Shen Yan, Xuehan Xiong, Arsha Nagrani, Anurag Arnab, Zhonghao Wang, Weina Ge,
  David Ross, and Cordelia Schmid.
\newblock Unloc: A unified framework for video localization tasks.
\newblock In \emph{Proceedings of the IEEE/CVF International Conference on
  Computer Vision}, pages 13623--13633, 2023.

\bibitem[Yang et~al.(2023)Yang, Chen, Zheng, Lu, and Wang]{yang2023basictad}
Min Yang, Guo Chen, Yin-Dong Zheng, Tong Lu, and Limin Wang.
\newblock Basictad: an astounding rgb-only baseline for temporal action
  detection.
\newblock \emph{Computer Vision and Image Understanding}, 232:\penalty0 103692,
  2023.

\bibitem[Yao et~al.(2024)Yao, Pi, Han, Liang, Xu, Zhang, Li, and
  Xu]{yao2024detclipv3}
Lewei Yao, Renjie Pi, Jianhua Han, Xiaodan Liang, Hang Xu, Wei Zhang, Zhenguo
  Li, and Dan Xu.
\newblock Detclipv3: Towards versatile generative open-vocabulary object
  detection.
\newblock In \emph{Proceedings of the IEEE/CVF Conference on Computer Vision
  and Pattern Recognition}, pages 27391--27401, 2024.

\bibitem[Yuan et~al.(2019)Yuan, Ma, Wang, Liu, and Zhu]{yuan2019semantic}
Yitian Yuan, Lin Ma, Jingwen Wang, Wei Liu, and Wenwu Zhu.
\newblock Semantic conditioned dynamic modulation for temporal sentence
  grounding in videos.
\newblock \emph{Advances in Neural Information Processing Systems}, 32, 2019.

\bibitem[Zeng et~al.(2019)Zeng, Huang, Tan, Rong, Zhao, Huang, and
  Gan]{zeng2019graph}
Runhao Zeng, Wenbing Huang, Mingkui Tan, Yu Rong, Peilin Zhao, Junzhou Huang,
  and Chuang Gan.
\newblock Graph convolutional networks for temporal action localization.
\newblock In \emph{Proceedings of the IEEE/CVF international conference on
  computer vision}, pages 7094--7103, 2019.

\bibitem[Zeng et~al.(2020)Zeng, Xu, Huang, Chen, Tan, and Gan]{zeng2020dense}
Runhao Zeng, Haoming Xu, Wenbing Huang, Peihao Chen, Mingkui Tan, and Chuang
  Gan.
\newblock Dense regression network for video grounding.
\newblock In \emph{Proceedings of the IEEE/CVF Conference on Computer Vision
  and Pattern Recognition}, pages 10287--10296, 2020.

\bibitem[Zeng et~al.(2024)Zeng, Zhong, Feng, and Ma]{zeng2024unimd}
Yingsen Zeng, Yujie Zhong, Chengjian Feng, and Lin Ma.
\newblock Unimd: Towards unifying moment retrieval and temporal action
  detection.
\newblock In \emph{European Conference on Computer Vision}, pages 286--304.
  Springer, 2024.

\bibitem[Zhang et~al.(2022)Zhang, Wu, and Li]{zhang2022actionformer}
Chen-Lin Zhang, Jianxin Wu, and Yin Li.
\newblock Actionformer: Localizing moments of actions with transformers.
\newblock In \emph{European Conference on Computer Vision}, pages 492--510.
  Springer, 2022.

\bibitem[Zhang et~al.(2023)Zhang, Li, and Bing]{zhang2023video}
Hang Zhang, Xin Li, and Lidong Bing.
\newblock Video-llama: An instruction-tuned audio-visual language model for
  video understanding.
\newblock \emph{arXiv preprint arXiv:2306.02858}, 2023.

\bibitem[Zhang et~al.(2020)Zhang, Peng, Fu, and Luo]{zhang2020learning}
Songyang Zhang, Houwen Peng, Jianlong Fu, and Jiebo Luo.
\newblock Learning 2d temporal adjacent networks for moment localization with
  natural language.
\newblock In \emph{Proceedings of the AAAI conference on artificial
  intelligence}, pages 12870--12877, 2020.

\bibitem[Zhao et~al.(2019)Zhao, Torralba, Torresani, and Yan]{zhao2019hacs}
Hang Zhao, Antonio Torralba, Lorenzo Torresani, and Zhicheng Yan.
\newblock Hacs: Human action clips and segments dataset for recognition and
  temporal localization.
\newblock In \emph{Proceedings of the IEEE/CVF International Conference on
  Computer Vision}, pages 8668--8678, 2019.

\bibitem[Zheng et~al.(2020)Zheng, Wang, Liu, Li, Ye, and
  Ren]{zheng2020distance}
Zhaohui Zheng, Ping Wang, Wei Liu, Jinze Li, Rongguang Ye, and Dongwei Ren.
\newblock Distance-iou loss: Faster and better learning for bounding box
  regression.
\newblock In \emph{Proceedings of the AAAI conference on artificial
  intelligence}, pages 12993--13000, 2020.

\bibitem[Zhu et~al.(2021)Zhu, Tang, Wang, Zheng, and Hua]{zhu2021enriching}
Zixin Zhu, Wei Tang, Le Wang, Nanning Zheng, and Gang Hua.
\newblock Enriching local and global contexts for temporal action localization.
\newblock In \emph{Proceedings of the IEEE/CVF international conference on
  computer vision}, pages 13516--13525, 2021.

\end{thebibliography}
}

\end{document}